\documentclass[conference]{IEEEtran}
\IEEEoverridecommandlockouts
\usepackage{times}
\usepackage{latexsym}
\usepackage{amsmath}
\usepackage{graphicx}
\usepackage{multirow}
\usepackage{url}
\usepackage{float}
\usepackage{stfloats}
\usepackage{amssymb}
\usepackage{array}
\usepackage{titlesec}
\usepackage{hyperref}
\usepackage{titlesec}

\titlespacing{\subsection}
  {0pt}    
  {6pt}    
  {4pt}    
\begin{document}
\title{ProtRLSearch: A Multi-Round Multimodal Protein Search Agent with Large Language Models Trained via Reinforcement Learning}

\def\BibTeX{{\rm B\kern-.05em{\sc i\kern-.025em b}\kern-.08em
		T\kern-.1667em\lower.7ex\hbox{E}\kern-.125emX}}

\author{
\IEEEauthorblockN{
Congying Liu$^{1}$,
Taihao Li$^{2}$,
Ming Huang$^{3}$,
Peipei Liu *$^{1,4}$\thanks{*Corresponding author: peipeiliu@yeah.net},
Xingyuan Wei$^{1,4}$,
Yiqing Shen$^{5}$,
Yanxu Mao$^{6}$,
Tiehan Cui$^{6}$
}

\IEEEauthorblockA{$^{1}$University of Chinese Academy of Sciences, Beijing, China}
\IEEEauthorblockA{$^{2}$Hangzhou Institute for Advanced Study, University of Chinese Academy of Sciences, Hangzhou, China}
\IEEEauthorblockA{$^{3}$Shenzhen Institutes of Advanced Technology, Chinese Academy of Sciences, Shenzhen, China}
\IEEEauthorblockA{$^{4}$Institute of Information Engineering, Chinese Academy of Sciences, Beijing, China}
\IEEEauthorblockA{$^{5}$Johns Hopkins University, Baltimore, USA}
\IEEEauthorblockA{$^{6}$Henan University, Kaifeng, China}
}

	\maketitle

\begin{abstract}
Protein analysis tasks arising in healthcare settings often require accurate reasoning under protein sequence constraints, involving tasks such as functional interpretation of disease-related variants, protein-level analysis for clinical research, and similar scenarios. To address such tasks, search agents are introduced to search protein-related information, providing support for disease-related variant analysis and protein function reasoning in protein-centric inference. However, such search agents are mostly limited to single-round, text-only modality search, which prevents the protein sequence modality from being incorporated as a multimodal input into the search decision-making process. Meanwhile, their reliance on reinforcement learning (RL) supervision that focuses solely on the final answer results in a lack of search process constraints, making deviations in keyword selection and reasoning directions difficult to identify and correct in a timely manner. To address these limitations, we propose ProtRLSearch, a multi-round protein search agent trained with multi-dimensional reward based RL, which jointly leverages protein sequence and text as multimodal inputs during real-time search to produce high quality reports. To evaluate the ability of models to integrate protein sequence information and text-based multimodal inputs in realistic protein query settings, we construct ProtMCQs, a benchmark of 3,000 multiple choice questions (MCQs) organized into three difficulty levels. The benchmark evaluates protein query tasks that range from sequence constrained reasoning about protein function and phenotype changes to comprehensive protein reasoning that integrates multi-dimensional sequence features with signal pathways and regulatory networks. Experimental results show that ProtRLSearch improves accuracy from 35.7\% to 86.9\% on level 1, from 30.5\% to 77.4\% on level 2, and from 26.1\% to 72.5\% on level 3. 

\end{abstract}
\section{Introduction}

In recent years, large language models (LLMs) have been introduced into the field of protein analysis to support variant effect interpretation and reasoning across protein and signaling pathway levels, among other protein-related analysis tasks, enabling disease-related variant analysis and protein reasoning in healthcare-related applications for clinical research~\cite{xiao2025proteinllm}. However, in the absence of direct access to real sequence information and relevant literature, LLMs often make them prone to hallucinations in tasks such as functional attribution and variant interpretation~\cite{huang2023survey}. As a result, reliable reasoning in protein analysis requires the integration of real-time online search, incorporating verifiable protein databases and literature as authentic protein search results to support protein multi-step reasoning by the model~\cite{apweiler2023uniprot}.
\begin{figure}[ht]
\centering
\includegraphics[width=\linewidth]{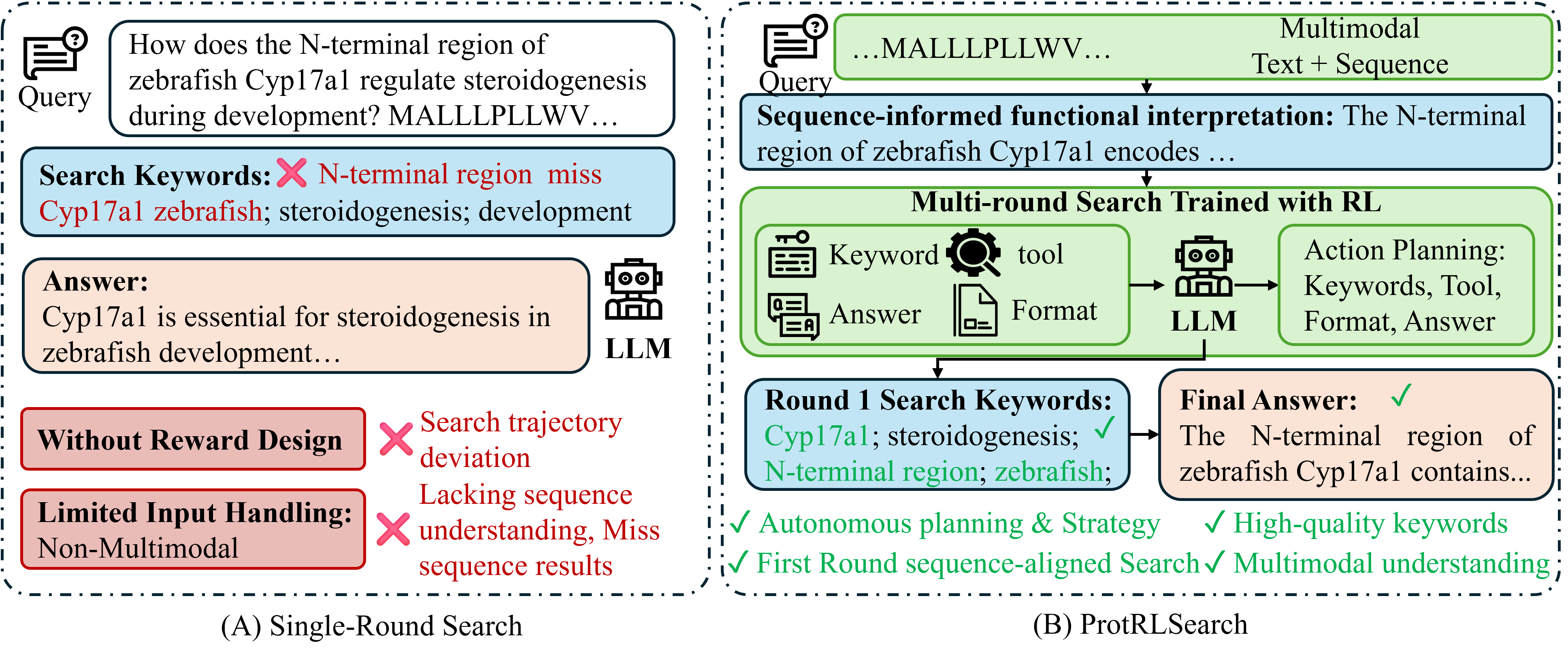}
\vspace{-8mm}
\caption{ A comparison between single-round search methods and the proposed approach, illustrating their differences in the search process. Red annotations indicate error cases such as missing keywords, while green annotations denote correct results.}
\vspace{-4mm}
\label{fig:case_study}
\end{figure}

Existing search agents such as BioMedSearch~\cite{liu2025biomedsearch} and MindSearch~\cite{chen2024mindsearch} primarily operate on the text modality and adopt a single-round search, where only one round of search from multiple databases is performed for a given query, followed by protein related reasoning based on the filtered results. Such search agents rely heavily on the keywords generated at the initial stage, while protein related conclusions are often strictly constrained by sequence level characteristics, including protein sequences, domains and mutation sites. When key sequence related information cannot be sufficiently expressed as textual keywords, models struggle to determine whether the search results are consistent with the given sequence, thereby limiting the effectiveness of sequence constrained reasoning. 

To address the limitations, researchers have proposed multi-round search agents such as Search-R1~\cite{jin2025search} and R1-Searcher~\cite{song2025r1}, which introduce reinforcement learning (RL) with final answer based rewards to regulate query generation and search round decisions, and progressively incorporate additional protein related textual information through multiple search rounds. However, the search planning of these methods still relies entirely on the text modality, and protein sequences are not incorporated as parallel inputs to constrain the search plan or guide decision making. As a result, multi-round search may still drift away from critical sequence features. In addition, these methods mainly depend on final answer rewards and impose limited constraints on search tool selection and search path planning, allowing errors in early search directions to persist and be amplified in subsequent rounds~\cite{huo2024multimodal}.

To address the limitations of prior work, we propose ProtRLSearch, a multi-round real-time search agent trained with RL using multi-dimensional reward signals, which takes protein sequences and textual information as multimodal inputs. Specifically, the agent leverages protein foundation models to learn representations of amino acid sequences, enabling the model to gain a deeper understanding of intrinsic properties related to protein structure and function. Through training with multi-dimensional reward signals based on keywords, answers, search tools, and output format, the search components including the Planner, Retriever and Executor learn to autonomously plan search processes and generate answers~\cite{jin2025search}. The Planner learns to extract representative keywords and assigns corresponding search tools by focusing on attributes involved in the queried protein sequence, such as interactions and variants, thereby defining subsequent search paths. The Retriever performs real-time searches over multiple databases to obtain candidate results and further filters the retrieved content based on multimodal queries, retaining information that matches the query for downstream reasoning. The Executor learns to perform reasoning and generates answers based on the search results and multimodal inputs, while assessing whether the current answers sufficiently support the requirements and thus deciding whether to trigger subsequent search rounds.

Existing datasets such as ProteinLMDataset~\cite{shen2024finetuning}, although incorporating multimodal information, lack explicit supervision over search and reasoning processes, which limits their applicability to protein search tasks. Therefore, we construct 3,000 multimodal training samples that explicitly associate multimodal inputs with search planning, answer information, and auxiliary fields. In addition, current benchmarks such as MedMCQA~\cite{pal2022medmcqa} do not evaluate whether models can answer protein-related questions by integrating sequence information with retrieval content~\cite{niu2025rethinking}. We construct ProtMCQs, a benchmark of 3,000 multiple choice questions (MCQs), covering tasks from sequence-constrained functional and phenotypic reasoning to integrative protein reasoning over multiple sequence features, signaling pathways, and regulatory networks. Our contributions are summarized as follows:
\begin{itemize}
    \item We propose ProtRLSearch, a multi-round multimodal protein search agent trained with RL. By integrating protein sequence representations into the search process, ProtRLSearch enables sequence-aware search decisions across multiple rounds and improves reasoning quality for protein-centric queries.
    \item We design a multi-dimensional reward scheme in the RL process, incorporating signals from keywords, format, search tools and answer quality. These reward components are efficiently integrated to guide the model in autonomously optimizing search paths.
    \item We introduce ProtMCQs, a benchmark dataset of 3,000 multimodal protein questions that spans three difficulty levels and diverse key protein scenarios for evaluating sequence-related reasoning ability.
    
\end{itemize}
\section{Related Works}
\subsection{Search Agent for Protein Query Tasks}
LLMs can generate explanations of domain functions and protein interactions in protein tasks~\cite{xiao2025proteinllm}. However, their outputs often exhibit inconsistencies with curated sequence annotations and structural knowledge~\cite{liu2025biomedsearch}, indicating the necessity of introducing search agents during reasoning to provide reliable protein information~\cite{gao2024rag}. Existing search agents, such as BioMedSearch~\cite{liu2025biomedsearch}, Pasa~\cite{he2024pasa} and MindSearch~\cite{chen2024mindsearch}, perform single-round search to integrate protein information from multiple sources, while Search-R1~\cite{jin2025search} and R1-Searcher~\cite{song2025r1} adopt multi-round search to progressively incorporate search results and refine answers. Nevertheless, these approaches often treat protein sequences as text and fail to sufficiently model the structural and functional semantics encoded in sequences, causing search paths to deviate from the stepwise structured process required for protein reasoning~\cite{zhang2024prot}. To address this limitation, ProtRLSearch integrates protein sequence and text as multimodal inputs, and performs multi-round search under sequence related constraints, enabling consistent alignment between search information and protein sequences.

\subsection{Reward Design in RL for Protein Tasks with LLMs}
RL has been applied to optimize search and reasoning in protein-related tasks through feedback signals~\cite{ouyang2022instructgpt}. BioReason~\cite{fallahpour2025bioreason} and Search-R1~\cite{jin2025search} introduce answer-based rewards in multi-step reasoning or multi-round search, improving alignment with reference answers and supporting the analysis of domain functions, protein interactions, and variant effects. However, these methods primarily rely on rewards defined on final answers and lack constraints on protein keywords identification and search tool selection, which limits the construction of effective search paths across heterogeneous data sources~\cite{orgad2024llms}. In contrast, ProtRLSearch adopts a multi-dimensional reward design to guide models in organizing multi-round online search around protein sequence, domain, and variant information~\cite{ranjan2024ragsurvey}.
\section{Methods}

\subsection{Multimodal Protein Sequence and Text Representation on LLM Backbone}

\begin{figure*}[ht]
\centering
\includegraphics[width=\textwidth,height=0.9\textheight,keepaspectratio]{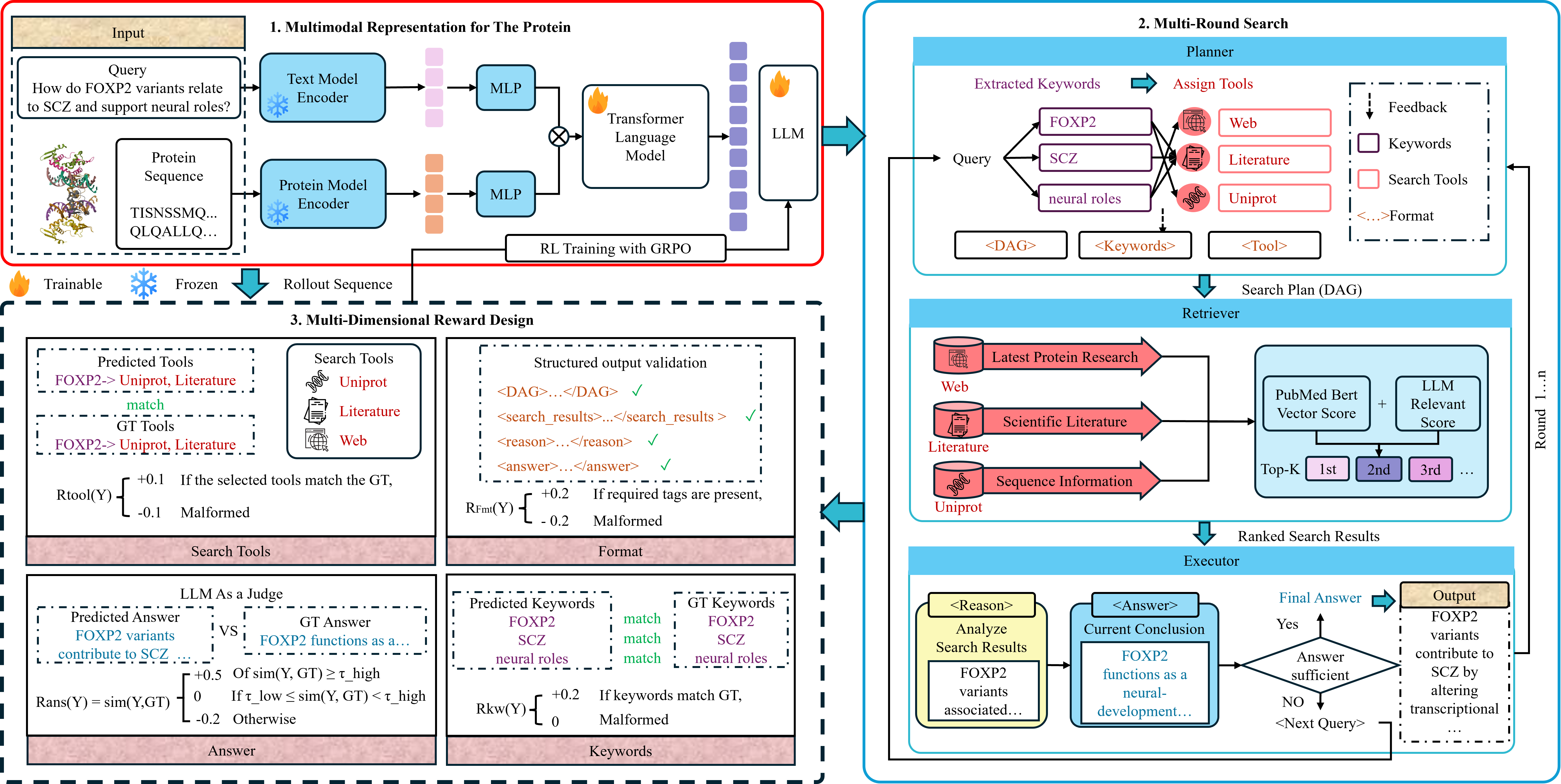}
\caption{The overall agent encompasses multimodal encoding, multi-round search trained via RL, and the design of multi-dimensional rewards.}
\vspace{-5mm}
\label{fig:methods}
\end{figure*}

We draw on the design philosophy of ProtLLM~\cite{zhuo2024protllm} and introduce a protein sequence representation module to extract feature representations from raw protein sequences. The model uses a frozen protein pretrained model such as ESM-2~\cite{lin2022language} to obtain residue-level contextual representations, covering domain boundaries, conserved motifs, and function-related sequence features. A learnable linear projection then maps these high-dimensional protein embeddings into the input space of LLM, enabling protein sequences to enter a unified transformer architecture together with textual information at the token level. To improve the alignment quality between sequences and semantics, we adopt the InterPT dataset released by ProtLLM~\cite{zhuo2024protllm} as the training corpus to guide the model in establishing stable mappings between protein regions and their corresponding descriptions. The projected protein embeddings are concatenated with the query text at the sequence level and jointly fed into LLM, allowing the model to process information from both modalities within a unified attention scope. As training progresses, the fusion layer gradually forms stable sequence and semantic alignment capability. The protein embeddings more accurately reflect residue segments, domain boundaries, and function-related signals in the LLM space, enabling the LLM to attend simultaneously to key regions in protein sequences and important descriptions in text. After training, the fused representations exhibit stronger cross-modal association and internal consistency as shown in Fig~\ref{fig:methods}, providing a coherent and interactive foundation for subsequent multi-round search.

\subsection{Multi-Round Search with Structured Outputs}
We construct a multi-round search agent composed of a Planner, a Retriever, and an Executor. The agent takes protein sequences and query as multimodal inputs and generates search plans, performs parallel multi-source search, and updates its conclusions.
 Unlike approaches that output only natural language answers, each round produces structured search outputs, including the search plan, ranked search results, and intermediate results used to generate the next-round query, supporting the progression of multi-round search.

In each search round, the Planner module uses an LLM to transform the current query into a search plan, as shown in Fig~\ref{fig:methods}. During this process, the model extracts multiple protein-centered keywords from the query and assigns search tools to each keyword, including Web for web search, Literature for document retrieval, and UniProt for protein database queries. The associations between keywords and tools are organized as a directed acyclic graph (DAG), which constitutes the structured output of the Planner for the current round and serves as the search plan for subsequent parallel searches.

The Retriever module conducts multi-source searches according to the DAG and obtains search results from web content, literature sources, and protein database annotations. The retrieved results are scored using two types of relevance signals. The first is a vector relevance score based on PubMedBERT~\cite{han2021pubmedbert}, which measures semantic similarity between the query and textual content. The second is an LLM-based relevance score, which evaluates the relevance of each search result under the given protein query. After combining the two relevance scores, the Retriever outputs a structured set of Top-K search results ranked by relevance for the current round.

The Executor module performs analysis based on the ranked retrieval results from the current round and outputs a set of structured result fields. Specifically, the model analyzes the retrieved content in \texttt{<reason>} and produces the intermediate conclusion for the current round in \texttt{<answer>}. The \texttt{<decide>} field indicates whether the available retrieval results are sufficient to address the original query, while the \texttt{<next\_query>} field provides the query required for the next search round when the results are insufficient. These structured outputs together form the analysis result of the current round, where \texttt{<next\_query>} is combined with the protein sequence as a new multimodal input in subsequent rounds to advance the next search and analysis process.

\subsection{Multi-Dimensional Reward  Design}

To guide LLMs to acquire stable search and reasoning capabilities in protein-related query tasks, we decompose the training objective into four verifiable sub objectives and construct a multi-dimensional reward function composed of Answer, Keywords, Tool, and Format. Each reward corresponds to an independent constraint illustrated in the Fig~\ref{fig:methods}. The overall reward is defined as a weighted sum of the sub rewards as follows:
\[R_{\text{total}} = \lambda_{\text{Ans}} R_{\text{Ans}} + \lambda_{\text{KW}} R_{\text{KW}} + \lambda_{\text{Tool}} R_{\text{Tool}} + \lambda_{\text{Fmt}} R_{\text{Fmt}} .\]
The Answer Reward is used to evaluate the consistency between the model generated answer and the ground truth (GT) answer as shown in Fig~\ref{fig:methods}. This reward is computed by an LLM-as-a-Judge, which scores the semantic similarity between the predicted answer and the GT answer. A positive reward is assigned when the similarity exceeds a predefined threshold, otherwise a negative reward is given. This reward constrains the model to produce final answers that align with the query objective after multi-round search; The Keywords Reward constrains the model behavior in extracting protein-related keywords during the first round of search planning. The predicted keywords set is matched against the keywords set derived from the GT. A positive reward is assigned when the predicted keywords match the GT keywords, while missing or mismatched keywords receive no reward or a penalty. This design emphasizes the central role of protein names, domains, and other core keywords in the search pipeline; The Tool Reward guides the model to select appropriate search tools based on the extracted keywords. The predicted keyword tool assignments are compared with the tool allocation defined in the GT. A positive reward is given when the predicted tools match the GT tools, while unreasonable tool selection or malformed outputs receive a penalty. This reward ensures that different types of protein-related keywords are mapped to correct data sources; The Format Reward constrains the model output to satisfy predefined structured format requirements. The output must contain complete and valid structured tags, such as \texttt{<DAG>}, \texttt{<search\_results>}, \texttt{<reason>}, and \texttt{<answer>}. A positive reward is assigned when all required tags are present and correctly formatted, otherwise a penalty is applied to ensure the parseability of the multi-round search.
\subsection{Training}
We train the model on a multimodal protein dataset of 3,000 samples via RL, where multi-dimensional rewards are derived from verifiable signals grounded in real-world protein databases and literature, thereby constraining cross-modal protein search without explicit supervision of intermediate search steps. Each sample is built upon a verifiable protein--species pair and validated against the UniProt reviewed database to ensure data reliability. The dataset covers six major protein categories, including transcription factors, protein regulators, ion channels and transporters, signaling and inflammation related proteins, synaptic and neuro developmental proteins, and mitochondrial proteins. It maintains a balanced distribution across four reasoning tasks, including variant to phenotype reasoning, structure to function reasoning, cross system mechanism reasoning, and cross species comparison. As shown in Fig~\ref{fig:train}, each sample is constructed through a multi-stage pipeline. First, a UniProt reviewed protein--species pair is selected and the corresponding protein sequence is retrieved. Next, function related literature is collected from PubMed. Finally, an LLM generates a research query, a DAG composed of keywords and search tools, and the associated reasoning tags \texttt{<reason>} and \texttt{<answer>}. This process organizes protein sequences, search results and related search information into structured training samples to support multi-round search.

\subsection{Benchmark Design}
\begin{figure}[ht]
\centering
\vspace{-5mm}
\includegraphics[width=\linewidth]{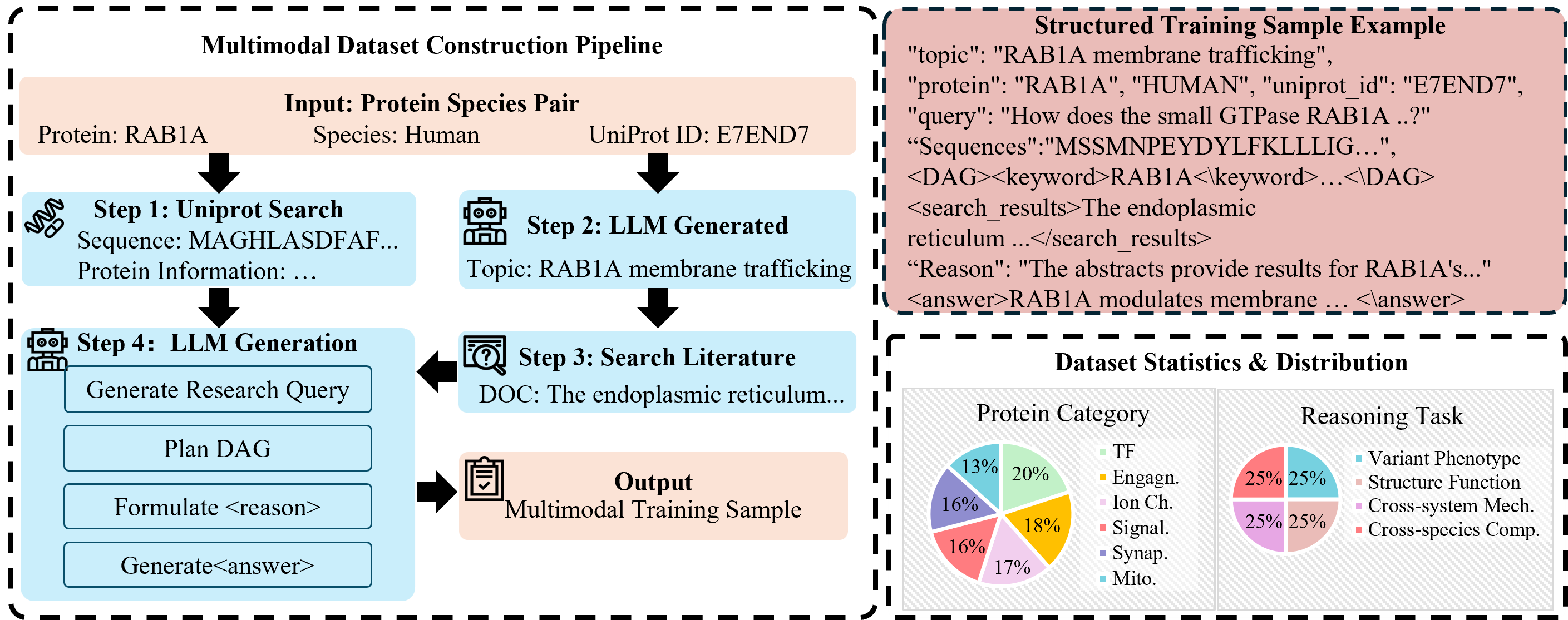}
\caption{Overview of the constructed training dataset}
\vspace{-5mm}
\label{fig:train}
\end{figure}
Existing biomedical datasets such as BioMedMCQs~\cite{liu2025biomedsearch} evaluate question answering performance in retrieval and information integration settings by simulating realistic user biomedical queries, whereas MedQA~\cite{jin2021disease} and MedMCQA~\cite{pal2022medmcqa} are primarily designed to measure knowledge understanding and text-based reasoning in medical examination and professional question answering scenarios. However, these datasets do not cover protein query settings in which biological sequences play a central role, thereby limiting the evaluation of protein-related capabilities of models~\cite{xu2022peer}. To address this gap, we construct ProtMCQs that jointly evaluates a model’s ability to understand protein sequences and to perform integrative reasoning over textual information during the search process. Its construction follows a workflow similar to a data generation pipeline, with detailed steps illustrated in Fig~\ref{fig:train}. ProtMCQs is organized into three progressively increasing difficulty levels to assess sequence-constrained reasoning in protein query tasks. Level 1 requires models, under the same sequence constraints, to integrate multiple relevant retrieved results around functional or phenotypic questions and to conduct joint reasoning over multiple sequence features in order to support judgments about functional changes or phenotypic differences. Level 2 introduces multiple distractor options generated based on existing studies involving sequence features, molecular properties, biological context, and species differences, requiring models to determine whether these options are consistent with the given sequence information. Level 3 further increases the reasoning complexity by requiring models to fuse multi-dimensional sequence features with information from signaling pathways and regulatory networks under a shared sequence context, in order to produce globally consistent protein-level reasoning decisions.

\section{EXPERIMENT}

\subsection{Implementation Details}
\begin{table*}[t]
\centering
\caption{Performance comparison across methods on BioMedMCQs and ProtMCQs.}
\label{tab:method_comparison}
\renewcommand{\arraystretch}{1.2}
\setlength{\tabcolsep}{4pt}
\begin{tabular}{
>{\centering\arraybackslash}m{1.6cm}
>{\centering\arraybackslash}m{2.6cm}
| c c c | c c c | c c c
}
\hline
\multirow{2}{*}{\textbf{Dataset}} 
& \multirow{2}{*}{\textbf{Method}} 
& \multicolumn{3}{c|}{\textbf{Level 1}}
& \multicolumn{3}{c|}{\textbf{Level 2}}
& \multicolumn{3}{c}{\textbf{Level 3}} \\
\cline{3-11}
& 
& \textbf{Acc\%} & \textbf{Token} & \textbf{Time/s}
& \textbf{Acc\%} & \textbf{Token} & \textbf{Time/s}
& \textbf{Acc\%} & \textbf{Token} & \textbf{Time/s} \\
\hline

\multirow{6}{*}{BioMedMCQs}
& Baseline
& 45.8{\tiny$\pm$0.6} & 321{\tiny$\pm$41} & 6.5{\tiny$\pm$0.7}
& 38.7{\tiny$\pm$1.2} & 335{\tiny$\pm$58} & 6.8{\tiny$\pm$1.1}
& 29.5{\tiny$\pm$0.9} & 379{\tiny$\pm$72} & 7.6{\tiny$\pm$1.4} \\

& BioMedSearch~\cite{liu2025biomedsearch}
& 73.3{\tiny$\pm$0.8} & 3029{\tiny$\pm$155} & 48.1{\tiny$\pm$1.5}
& 69.7{\tiny$\pm$1.4} & 3348{\tiny$\pm$121} & 65.1{\tiny$\pm$0.9}
& 62.4{\tiny$\pm$1.0} & 3987{\tiny$\pm$188} & 78.9{\tiny$\pm$1.6} \\

& BioReason~\cite{fallahpour2025bioreason}
& 63.9{\tiny$\pm$1.7} & 472{\tiny$\pm$63} & 11.7{\tiny$\pm$1.3}
& 56.2{\tiny$\pm$0.8} & 496{\tiny$\pm$71} & 16.7{\tiny$\pm$0.8}
& 49.3{\tiny$\pm$1.4} & 528{\tiny$\pm$53} & 20.5{\tiny$\pm$0.6} \\

& Search-R1~\cite{jin2025search}
& 76.1{\tiny$\pm$0.7} & 2063{\tiny$\pm$134} & 19.3{\tiny$\pm$0.4}
& 70.9{\tiny$\pm$1.3} & 2582{\tiny$\pm$98}  & 27.1{\tiny$\pm$1.2}
& 63.5{\tiny$\pm$1.8} & 2835{\tiny$\pm$176} & 32.4{\tiny$\pm$1.9} \\

& ProtLLM~\cite{zhuo2024protllm}
& 74.8{\tiny$\pm$1.1} & 409{\tiny$\pm$82}  & 9.6{\tiny$\pm$1.1}
& 60.4{\tiny$\pm$0.9} & 534{\tiny$\pm$65}  & 14.9{\tiny$\pm$1.3}
& 57.9{\tiny$\pm$0.5} & 560{\tiny$\pm$103} & 16.2{\tiny$\pm$1.5} \\

& ProtRLSearch(Ours)
& \textbf{89.2{\tiny$\pm$1.4}} & 554{\tiny$\pm$117} & 10.5{\tiny$\pm$0.9}
& \textbf{75.8{\tiny$\pm$0.6}} & 785{\tiny$\pm$84}  & 13.4{\tiny$\pm$1.6}
& \textbf{71.7{\tiny$\pm$1.9}} & 873{\tiny$\pm$156} & 21.3{\tiny$\pm$1.7} \\
\hline

\multirow{6}{*}{ProtMCQs}
& Baseline
& 35.7{\tiny$\pm$0.8} & 432{\tiny$\pm$74} & 8.9{\tiny$\pm$0.9}
& 30.5{\tiny$\pm$1.4} & 468{\tiny$\pm$92} & 9.6{\tiny$\pm$1.2}
& 26.1{\tiny$\pm$1.1} & 493{\tiny$\pm$63} & 12.4{\tiny$\pm$0.6} \\

& BioMedSearch~\cite{liu2025biomedsearch}
& 61.4{\tiny$\pm$0.9} & 3217{\tiny$\pm$188} & 53.8{\tiny$\pm$1.3}
& 57.3{\tiny$\pm$1.8} & 3129{\tiny$\pm$146} & 58.0{\tiny$\pm$0.7}
& 48.1{\tiny$\pm$0.7} & 3572{\tiny$\pm$171} & 69.1{\tiny$\pm$2.5} \\

& BioReason~\cite{fallahpour2025bioreason}
& 58.9{\tiny$\pm$1.6} & 365{\tiny$\pm$55} & 12.6{\tiny$\pm$1.1}
& 50.6{\tiny$\pm$0.6} & 503{\tiny$\pm$73} & 13.3{\tiny$\pm$0.8}
& 47.4{\tiny$\pm$1.3} & 514{\tiny$\pm$97} & 15.9{\tiny$\pm$1.4} \\

& Search-R1~\cite{jin2025search}
& 65.6{\tiny$\pm$0.5} & 2376{\tiny$\pm$162} & 19.3{\tiny$\pm$2.6}
& 58.1{\tiny$\pm$1.9} & 2582{\tiny$\pm$183} & 27.6{\tiny$\pm$1.3}
& 49.2{\tiny$\pm$1.2} & 2935{\tiny$\pm$115} & 32.4{\tiny$\pm$1.1} \\

& ProtLLM~\cite{zhuo2024protllm}
& 72.1{\tiny$\pm$1.4} & 448{\tiny$\pm$69}  & 10.5{\tiny$\pm$1.5}
& 62.8{\tiny$\pm$0.7} & 514{\tiny$\pm$82}  & 16.3{\tiny$\pm$1.4}
& 59.1{\tiny$\pm$1.0} & 546{\tiny$\pm$104} & 18.2{\tiny$\pm$0.9} \\

& ProtRLSearch(Ours)
& \textbf{86.9{\tiny$\pm$1.3}} & 632{\tiny$\pm$118} & 15.9{\tiny$\pm$0.7}
& \textbf{77.4{\tiny$\pm$0.8}} & 706{\tiny$\pm$91}  & 20.4{\tiny$\pm$1.2}
& \textbf{72.5{\tiny$\pm$1.7}} & 989{\tiny$\pm$153} & 24.3{\tiny$\pm$0.4} \\
\hline
\end{tabular}
\vspace{-5mm}
\end{table*}

On the BioMedMCQs~\cite{liu2025biomedsearch} and ProtMCQs multiple-choice datasets, we record model accuracy, the number of generated tokens, and the average inference time. All models adopt a unified input format and evaluation template to ensure the comparability of results. In addition, we evaluate the compared methods on 1000 randomly sampled questions from MedMCQA~\cite{pal2022medmcqa} and MedQA~\cite{jin2021disease} to assess generalization to general medical question answering scenarios beyond protein-related tasks. All comparison methods are based on Qwen3-8B~\cite{yang2025qwen3} with identical model configurations. For all search methods, the number of retrieved passages is uniformly set to~3. The weight configuration of the multi-dimensional reward signals in ProtRLSearch is defined as:
\[
\lambda_{\text{Ans}} = 0.5,\quad
\lambda_{\text{KW}} = 0.2,\quad
\lambda_{\text{Tool}} = 0.2,\quad
\lambda_{\text{Fmt}} = 0.1.
\]
The answer reward is assigned the largest weight to prioritize alignment with the task objective, the keywords and search tool rewards share equal weights to jointly constrain search direction and tool selection, and the format reward serves as an auxiliary constraint to ensure output regularity.
\begin{table*}[t]
\caption{Performance comparison across BioMedMCQs and ProtMCQs under different pre-training configurations.}
\label{tab:pretrain_ablation}
\centering
\scriptsize
\setlength{\tabcolsep}{3pt}
\resizebox{\textwidth}{!}{
\begin{tabular}{
>{\centering\arraybackslash}m{2.2cm}
>{\centering\arraybackslash}m{1.2cm}
>{\centering\arraybackslash}m{0.8cm}
| c c c | c c c | c c c
}
\hline
\multirow{2}{*}{\textbf{Dataset}} 
& \multirow{2}{*}{\textbf{Protein}} 
& \multirow{2}{*}{\textbf{RL}}
& \multicolumn{3}{c|}{\textbf{Level 1}} 
& \multicolumn{3}{c|}{\textbf{Level 2}}
& \multicolumn{3}{c}{\textbf{Level 3}} \\
\cline{4-12}
& & 
& \textbf{Acc\%} & \textbf{Token} & \textbf{Time/s}
& \textbf{Acc\%} & \textbf{Token} & \textbf{Time/s}
& \textbf{Acc\%} & \textbf{Token} & \textbf{Time/s} \\
\hline

\multirow{3}{*}{BioMedMCQs}
& $\checkmark$ & $\checkmark$
& \textbf{89.2{\tiny$\pm$1.4}} & \textbf{554{\tiny$\pm$117}} & \textbf{10.5{\tiny$\pm$0.9}}
& \textbf{75.8{\tiny$\pm$0.6}}  & \textbf{785{\tiny$\pm$84}} & \textbf{13.4{\tiny$\pm$1.6}}
& \textbf{71.7{\tiny$\pm$1.9}} & \textbf{873{\tiny$\pm$156}} & \textbf{21.3{\tiny$\pm$1.7}} \\

& $\checkmark$ & $\times$
& 76.7{\tiny$\pm$1.9}  & 977{\tiny$\pm$148} & 15.4{\tiny$\pm$2.3}
& 70.7{\tiny$\pm$0.9} & 1719{\tiny$\pm$238} & 20.5{\tiny$\pm$1.7}
& 65.1{\tiny$\pm$1.8} & 2178{\tiny$\pm$327} & 35.6{\tiny$\pm$1.3} \\

& $\times$ & $\checkmark$
& 83.2{\tiny$\pm$0.6} & 586{\tiny$\pm$103} & 12.6{\tiny$\pm$1.2}
& 72.9{\tiny$\pm$1.5} & 874{\tiny$\pm$129} & 15.3{\tiny$\pm$2.1}
& 69.8{\tiny$\pm$0.7} & 910{\tiny$\pm$203} & 22.1{\tiny$\pm$1.8} \\
\hline

\multirow{3}{*}{ProtMCQs}
& $\checkmark$ & $\checkmark$
& \textbf{86.9{\tiny$\pm$1.3}} & \textbf{632{\tiny$\pm$118}} & \textbf{15.9{\tiny$\pm$0.7}}
& \textbf{77.4{\tiny$\pm$0.8}}  & \textbf{706{\tiny$\pm$91}} & \textbf{20.4{\tiny$\pm$1.2}}
& \textbf{72.5{\tiny$\pm$1.7}} & \textbf{989{\tiny$\pm$153}} & \textbf{24.3{\tiny$\pm$0.4}} \\

& $\checkmark$ & $\times$
& 78.6{\tiny$\pm$1.7} & 1016{\tiny$\pm$175} & 24.7{\tiny$\pm$1.8}
& 63.7{\tiny$\pm$1.0} & 1623{\tiny$\pm$136} & 22.8{\tiny$\pm$1.3}
& 61.2{\tiny$\pm$1.4} & 2031{\tiny$\pm$164} & 33.4{\tiny$\pm$2.0} \\

& $\times$ & $\checkmark$
& 73.5{\tiny$\pm$0.5} &  795{\tiny$\pm$134} & 19.2{\tiny$\pm$2.5}
& 61.2{\tiny$\pm$1.9} & 1072{\tiny$\pm$148} & 25.7{\tiny$\pm$1.9}
& 55.4{\tiny$\pm$1.1} & 1518{\tiny$\pm$172} & 29.7{\tiny$\pm$1.5} \\
\hline
\end{tabular}
}
\vspace{-5mm}
\end{table*}

\subsection{Main Results}

\begin{table}[t]
\centering
\caption{Performance comparison on MedMCQA and MedQA (Acc
\%).}
\label{tab:medqa_comparison}

\begin{tabular}{l|c c}
\hline
\textbf{Method} & \textbf{MedMCQA} & \textbf{MedQA} \\
\hline
BioBERT~\cite{lee2020biobert}                 & 36.7 & 37.1 \\
SciBERT~\cite{beltagy2019scibert}                 &   -  & 39.2 \\
MedAlpaca-7B~\cite{han2023medalpaca}            & 55.2 & 45.8 \\
MEDAGENTS~\cite{tang2024medagents}               & 64.1 & 59.3 \\
\hline
ProtRLSearch (w/o RL)   & 85.8 & 83.9 \\
ProtRLSearch (w/o Prot) & 88.5 & 85.2 \\
ProtRLSearch            & \textbf{90.4} & \textbf{87.4} \\
\hline
\end{tabular}
\vspace{-8mm}
\end{table}

Table~\ref{tab:method_comparison} reports the overall performance of ProtRLSearch and competing methods on BioMedMCQs~\cite{liu2025biomedsearch} and ProtMCQs. Across all difficulty levels on both datasets, ProtRLSearch achieves the highest accuracy while maintaining controllable, and in some cases superior, inference cost. On BioMedMCQs~\cite{liu2025biomedsearch}, ProtRLSearch attains 89.2\%, 75.8\%, and 71.7\% accuracy at levels 1, 2, and 3, respectively, representing substantial improvements over the baseline results of 45.8\%, 38.7\%, and 29.5\%. Meanwhile, its average inference times are 10.5s, 13.4s and 21.3s, only slightly higher than the baseline single-step inference time and without exponential growth as task difficulty increases, indicating that introducing multi-round search does not incur notable efficiency loss. In contrast, Search-R1~\cite{jin2025search} achieves accuracies of 76.1\%, 70.9\% and 63.5\% at the corresponding levels, but with substantially longer inference times. These results indicate that, by incorporating multimodal information and multi-dimensional reward signals, ProtRLSearch enables multi-round search to converge earlier to effective retrieval paths and avoids unnecessary inference overhead. On ProtMCQs, ProtRLSearch achieves 86.9\%, 77.4\% and 72.5\% accuracy at level 1–3, again substantially outperforming the baseline results of 35.7\%, 30.5\% and 26.1\%, highlighting the critical role of introducing search and planning under sequence constraints. In terms of efficiency, the corresponding inference times are 15.9s, 20.4s and 24.3s. Although higher than the baseline single-step inference time, this increase is acceptable relative to the magnitude of the performance gains. ProtLLM~\cite{zhuo2024protllm} benefits from protein sequence modeling and outperforms BioReason on this dataset; however, due to the lack of search planning, it still falls notably behind ProtRLSearch on high-difficulty queries that require cross-region reasoning with retrieved information. Overall, these results further demonstrate that multi-round search planning based on multimodal protein sequence encodings can achieve a favorable balance between accuracy improvement and efficiency control.

Table ~\ref{tab:medqa_comparison} reports the performance comparison on MedMCQA~\cite{pal2022medmcqa} and MedQA~\cite{jin2021disease}. ProtRLSearch achieves the highest accuracy on both datasets, reaching 90.4\% on MedMCQA~\cite{pal2022medmcqa} and 87.4\% on MedQA~\cite{jin2021disease}, outperforming existing biomedical language models and agent-based baseline. These results indicate that the proposed method exhibits stronger reasoning stability and generalization ability in general medical question answering scenarios.

\subsection{Ablation Study} 
To analyze the independent contributions of RL and multimodal modeling in ProtRLSearch, we conduct ablation experiments on three difficulty levels of the BioMedMCQs~\cite{liu2025biomedsearch} and ProtMCQs datasets in table~\ref{tab:pretrain_ablation}. We construct w/o Protein, which removes the protein modality, and w/o RL, which removes RL, while the full setting includes both components. As shown in table~\ref{tab:pretrain_ablation}, the full setting achieves the highest accuracy across all difficulty levels on both datasets, together with lower Token consumption and retrieval inference time. On BioMedMCQs~\cite{liu2025biomedsearch}, removing RL reduces the level 1 accuracy from 89.2\% to 76.7\% and increases Token and time cost; in contrast, removing the protein modality leads to a limited performance drop, indicating that in biomedical pure-text tasks, protein sequences are not a determining factor for search and reasoning performance. On ProtMCQs, removing the protein modality causes the largest performance degradation, with level 3 accuracy decreasing from 72.5\% to 55.4\%, together with higher Token consumption and inference time; in contrast, w/o RL shows lower accuracy than the full setting across all difficulty levels and is associated with higher inference cost. Overall, the full setting outperforms all ablation variants across both datasets and all difficulty levels, indicating that introducing RL and multimodal modeling supports more stable overall performance. The ablation results in table~\ref{tab:medqa_comparison} show that removing either the RL component or the protein-related modeling leads to consistent accuracy degradation on both MedMCQA~\cite{pal2022medmcqa} and MedQA~\cite{jin2021disease}, which may be attributed to the impact on search decision making, thereby reducing overall answer stability.

\section{Conclusion} 
In protein-related generation and reasoning tasks, LLMs often produce unstable or factually inconsistent outputs due to the lack of explicit modeling of protein sequence information and real-time retrieval capabilities, particularly in healthcare-oriented scenarios such as disease-related variant analysis. To address this limitation, we propose ProtRLSearch, a multi-round multimodal protein search agent trained via RL. ProtRLSearch jointly models protein sequence representations and textual information, incorporates multi-round search with LLMs, and learns stable search trajectories within a closed-loop framework composed of a Planner, a Retriever, and an Executor under multi-dimensional reward constraints. Experimental results indicate that explicitly treating protein sequences as a complementary modality alongside text contributes to more stable and consistent multi-round protein search and reasoning. Future work will explore adaptive protein representation learning and online search supervision to improve generalization to unseen protein families and dynamically evolving knowledge sources.
	\bibliographystyle{IEEEtran}     
	\bibliography{references_main}        
\end{document}